\documentclass{ecai} 

\usepackage{latexsym}
\usepackage{amssymb}
\usepackage{amsmath}
\usepackage{amsthm}
\usepackage{booktabs}
\usepackage{enumitem}
\usepackage{graphicx}
\usepackage{color}

\usepackage{times}
\usepackage{soul}
\usepackage{url}
\usepackage[hidelinks]{hyperref}
\usepackage[utf8]{inputenc}
\usepackage[small]{caption}
\usepackage{graphicx}
\usepackage{amsmath}
\usepackage{amsthm}
\usepackage{booktabs}
\usepackage{algorithm}
\usepackage{algorithmic}
\usepackage[switch]{lineno}

\usepackage{amsfonts} 
\usepackage{balance}
\usepackage{todonotes}
\usepackage{nicefrac}
\usepackage{natbib}

\newcommand{\BibTeX}{B\kern-.05em{\sc i\kern-.025em b}\kern-.08em\TeX}

\usepackage{xcolor}         

\begin{document}


\begin{frontmatter}


\paperid{123} 


\title{Counterfactual Explanations for Clustering Models}


\author[A]{\fnms{Aurora}~\snm{Spagnol}}
\author[B]{\fnms{Kacper}~\snm{Sokol}\orcid{0000-0002-9869-5896}}
\author[A]{\fnms{Pietro}~\snm{Barbiero}\orcid{0000-0003-3155-2564}} 
\author[A]{\fnms{Marc}~\snm{Langheinrich}\orcid{0000-0002-8834-7388}} 
\author[A]{\fnms{Martin}~\snm{Gjoreski}\orcid{0000-0002-1220-7418}\thanks{Corresponding Author. Email: martin.gjoreski@usi.ch.}}

\address[A]{Università della Svizzera italiana (USI), Lugano}
\address[B]{ETH Zurich, Zurich}


\begin{abstract}
    Clustering algorithms rely on complex optimisation processes that may be difficult to comprehend, especially for individuals who lack technical expertise.
    While many explainable artificial intelligence techniques exist for supervised machine learning, unsupervised learning -- and clustering in particular -- has been largely neglected.
    To complicate matters further, the notion of a ``true'' cluster is inherently challenging to define.
    These facets of unsupervised learning and its explainability make it difficult to foster trust in such methods and curtail their adoption.
    To address these challenges, we propose a new, model-agnostic technique for explaining clustering algorithms with counterfactual statements. 
    Our approach relies on a novel soft-scoring method that captures the spatial information utilised by clustering models. 
    It builds upon a state-of-the-art Bayesian counterfactual generator for supervised learning to deliver high-quality explanations. %
    We evaluate its performance on five datasets and two clustering algorithms, and demonstrate that
    introducing soft scores to guide counterfactual search significantly improves the results.
\end{abstract}
\end{frontmatter}


\section{Introduction}

Clustering plays a vital role in artificial intelligence (AI) and machine learning (ML), allowing us to discover hidden patterns and structures in data when labels are missing~\cite{ezugwu2022comprehensive}. However, as clustering techniques evolved, they have become more opaque, posing challenges for humans to understand the rationale behind cluster assignments, thus undermining user trust in the outcomes~\cite{rudin2019stop}. 
In recent years, the concept of eXplainable AI (XAI) has gained significant traction as a vital bridge between complex ML models and human understanding~\cite{sokol2021explainability}. 

XAI has notably found numerous applications in supervised learning and some in reinforcement learning~\cite{Ali2023}, but its adoption in unsupervised learning, and clustering more specifically, has been overlooked. In the context of clustering, interpretable (transparent) methods rely on tree models or recasting classic clustering algorithms like $k$-means into neural networks. However, these techniques may sacrifice flexibility, accuracy and usability due to their strong limits on model selection, explanation and representation. On the other hand, post-hoc explainable techniques offer greater flexibility with fewer underlying assumptions~\cite{RibeiroSingh2016}.
Existing post-hoc approaches primarily focus on identifying associations between input features and cluster assignments~\cite{Louhichi2023,Altmann2010,ellis2021algorithm,scholbeck2023algorithm}, but these associations lack actionable insights~\cite{Lundberg2017,Lakkaraju2016}.
This gap in the XAI landscape highlights the need for novel techniques that generate actionable explanations tailored to clustering models.

%
Counterfactual explanations 
enable users to explore alternative ``what-if'' scenarios.
Similar to counterfactuals for supervised~\cite{romashov2022baycon,Dandl2020,poyiadzi2020face} and reinforcement~\cite{gajcin2023rl,olson2021counterfactual} learning, counterfactuals for clustering can show how input modifications might alter cluster assignments. 
For instance, in a scenario where clustering is used for customer segmentation, a customer who wants to receive premium offers, can use counterfactual explanations to determine what specific actions or changes in behavior (e.g., buying wine from the company's store instead of a local winery) would make a company offer them a premium membership.

Unlike supervised and reinforcement learning, where individual instances or agent actions can be probed, clustering relies on the entire dataset to construct the model and assign labels. However, access to the entire dataset for explanation generation should be avoided for privacy reasons, hence only a few descriptive instances and cluster statistics should be provided. This is also important for large datasets that are impossible to process in their entirety. 
Post-hoc XAI is an attractive paradigm for explaining clustering models since it does not require access to the entire dataset and allows transparency without sacrificing privacy.
The nature of clustering -- assigning collections of data points to clusters -- presents unique challenges, as the vast majority of algorithms is \emph{transductive}, i.e., they do not allow assigning cluster memberships to new, unseen instances without refitting the entire model. Moreover, finding an approach to capture the spatial information upon which the clustering is built -- distances, densities, and other relations between instances and the data space -- is non-trivial.

To address these challenges, we designed a novel \emph{soft-scoring} technique to capture the spatial information that clustering models rely upon, which we integrated into a state-of-the-art Bayesian counterfactual generator -- BayCon~\cite{romashov2022baycon}.
%
%
%
Through this connection, we propose a novel XAI method specifically designed for clustering, which 
addresses the lack of \textbf{model-agnostic} tools that are able to provide \textbf{actionable} counterfactual explanations for clustering models.
Our method, based on a bespoke soft-scoring technique, consistently outperforms the hard-scoring baseline across all experiments in terms of the percentage of instances explained, and on some datasets, it even outperforms the model-specific soft-scoring techniques, while maintaining similar computational complexity (e.g., execution time). 
Similarly, the distance between the initial instance ($x^\star$) and candidate counterfactual ($x^\prime$) remains almost unaffected ($S_{X}$ in Equation~\ref{eq:1}), as well as the number of feature-tweaks between the two ($S_{f}$ in Equation~\ref{eq:1}). 
These results showcase the method's ability to generate high-quality counterfactuals, even in scenarios where the baseline approach could not find any counterfactual explanations.

\section{Background and Problem statement}
\subsection{Clustering Taxonomy} \label{clustering}
When trying to derive insights through clustering, it is common practice to experiment with different algorithms and compare assignments relying on quantitative scores and visualisations. Thus, the proposed explainability method should support as many clustering algorithms as possible. For the reasons anticipated, we chose to design a post-hoc method, which implies the possibility of relying on model agnosticism. 
To make sure we are fully complying with this feature, it is necessary to explore clustering taxonomies, identify the most common families of algorithms, and verify if the proposed method can potentially be applied to all of them. Typically, clustering algorithms are classified into three broad types: partitional, hierarchical, and density-based. 

Recently, more nuanced taxonomies have emerged, proposing additional categories such as grid-based and model-based algorithms~\cite{Fahad2014}. Moreover, clustering techniques have expanded beyond traditional categories and incorporated new concepts from diverse domains like fuzzy theory, graph theory, fractal theory~\cite{Xu2015}, and, more recently, deep learning~\cite{zhou2022comprehensive}. 

However, the emergence of hybrid techniques has blurred the boundaries between these categories. For instance, the Hierarchical Density-Based Spatial Clustering of Applications with Noise (HDBSCAN~\cite{campello2013density}) and the Growing Hierarchical EXIN (GH-EXIN~\cite{cirrincione2020gh}) combine the advantages of hierarchical and density-based approaches. Clustering In QUEst (CLIQUE~\cite{agrawal1998automatic}) and STatistical INformation Grid (STING~\cite{wang1997sting}), though initially simply classified as grid-based algorithms, also include elements of hierarchical and density-based methods, respectively.

While some of the algorithm categories named are not suitable for tabular data, we tried to represent all the applicable ones, starting from the clustering methods available in scikit-learn.

\subsection{Transduction and Agnosticism in Clustering}
Clustering is an unsupervised learning task with the goal of grouping similar data points together based on a given similarity metric. Finding natural groups or clusters in the data is desirable as it may be useful for additional data analysis or interpretation. However, due to several variables (further discussed in the taxonomy presented in Section~\ref{clustering}), including noise, outliers and high-dimensionality of data, detecting actual clusters in real-world datasets is inherently difficult~\cite{Jain1999}. The complexity of assessing clustering goodness makes this task a perfect candidate for the application of XAI techniques.

We set out to design a model-agnostic XAI method that works with various clustering algorithms, which is a challenging task given that most such models are \emph{transductive}, i.e., once the clusters are built, the algorithms cannot ``predict'' cluster membership of new data points without refitting the models. To avoid this constraint, many clustering algorithms also exploit the $k$-nearest neighbours ($k$-NN) classifier to assign membership of new instances to clusters, which are represented by their respective medoids. However, we avoided these implementations to avoid forcing unfounded cluster assignments. 

Typically, clustering algorithms are classified into three broad types: partitional, hierarchical and density-based (Section ~\ref{clustering} offers a more detailed discussion). Our experimental setup included  HDBSCAN -- a representative of hierarchical and density-based (non-parametric) algorithms -- and $k$-means -- a representative of non-hierarchical and centroid-based (parametric) algorithms. GMMs are another non-hierarchical and centroid-based that can be easily used instead of $k$-means. Furthermore,  $k$-means is a transductive algorithm because assigning a cluster to a new data point is equivalent to performing the Expectation (\emph{E}) step of the algorithm, i.e., the stage in which data points are assigned to the closest centroid. In DBSCAN~\cite{ester1996density}, the distance metric used by the algorithm can be utilised to check if the new data point is close enough to one of the cluster \emph{core samples} or if it is an outlier. 

\subsection{Problem statement}
To assess the quality of the counterfactuals ($x^\prime$) with respect to the initial instance ($x^\star$) and the trained clustering model, we use the following scoring function used as part of optimisation maximisation problem, i.e., the higher the score, the better the counterfactual:
\begin{equation}
F(x^\prime, x^\star) = S_f \cdot S_x \cdot S_y 
\text{~,}
\label{eq:1}
\end{equation}
where $S_{f}$ is the number of non-altered features, $S_{x}$ is the similarity in the feature space, and $S_{y}$ is the similarity in the prediction space , all scaled in to the $[0, 1]$ range. The main novelty of this study is the soft-scoring technique for calculating $S_{y}$. The other two components, $S_{x}$ and $S_{f}$, are used as defined by BayCon: 
\begin{equation*}
S_f(x^\prime, x^\star) = \frac{\text{\# of equal features between } x^\prime \text{ and } x^\star} {\text{Overall \# of features}}
\end{equation*}
and
\begin{equation*}
S_x(x^\prime, x^\star) = 1 - d_{\text{Gower}}
\text{~,}
\end{equation*}
where, $d_{\text{Gower}}$ is the Gower distance -- a metric that can operate in mixed (numerical and categorical) feature spaces. 

\begin{figure}[t]
    \centering
    \includegraphics[width=\linewidth]{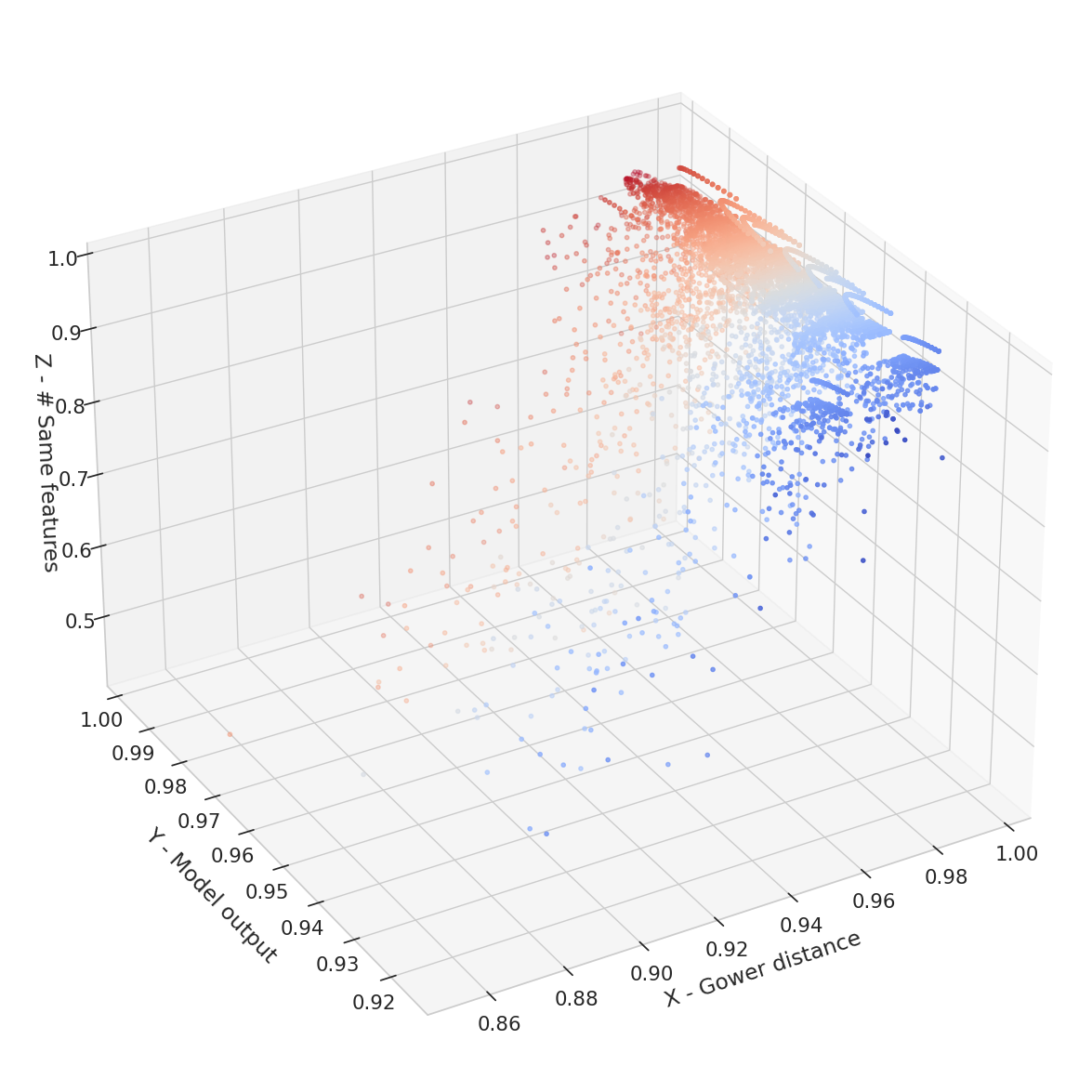}
    \caption[$k$-means++ soft scores visualisation]{Candidate counterfactuals for a $k$-means++ clustering model built on the Biodeg dataset. The colour-coding captures the quality of the counterfactuals from blue -- worse -- to red -- better. x-axis represents Gower distance; y-axis shows distance in the prediction space; and z-axis is the proportion of unchanged features (the number of features that were not tweaked divided by the total number of features).\label{Objective_Scores}}
    \vspace*{6ex}
\end{figure}

Figure~\ref{Objective_Scores} depicts the importance of cluster-specific soft-scores in the prediction space $S_{y}$, which allows us to rank candidate explanations with more granularity. More specifically, the figure depicts example optimisation soft scores for candidate counterfactuals, which are colour-coded, from worse (blue) to better (red).
The x-axis represents the Gower distance ($1-S_{x}$); the y-axis represents the soft-scores in the prediction space ($S_{y}$); and the z-axis represents the distance in the number of changed features ($1/S_{f}$).
The optimisation score is the highest for (i) the counterfactuals that are closest to the centroid of the target cluster (y-axis); (ii) have lower Gower distance from the explained instance; and (iii) require fewer features to be tweaked.
Note that ranking based on the prediction score would not be possible if we simply applied any supervised counterfactual generator (e.g., BayCon) because such an approach would provide only binary information on whether the cluster assignment has been flipped to a predefined one or not.

\section{Method}
In this section, we briefly introduce BayCon -- a Baesian counterfactual generator for supervised models -- which our method builds upon~\cite{romashov2022baycon}. We adapt BayCon to clustering algorithms by introducing two \textbf{model-specific} soft scores: one for $k$-means (non-hierarchical, centroid-based algorithm) utilising scaled distances from the centroid of the target cluster, and another for HDBSCAN (hierarchical, density-based algorithm) utilising probabilities of cluster membership provided by the method~\cite{mcinnes2017hdbscan}. 
Then, we introduce a novel \textbf{model-agnostic} soft-scoring method. Along the lines of the model-specific soft scoring, we first extract representative points for each cluster using prototypes and criticisms~\cite{Kim2016ExamplesAN}, which are a global and model-agnostic method for XAI. Subsequently, the selected data instances are employed to train a semi-supervised self-training classifier, which assigns membership probabilities based on the clusters.  

\subsection{Counterfactual Search with Hard Scores}
BayCon has several characteristics desirable of counterfactual explainers: time efficiency, scalability, model-agnosticism (for supervised models), and compatibility with continuous and categorical features. It is optimised for feature sparsity (i.e., the explanations require few feature tweaks), and avoids out-of-distribution explanations by using an outlier detection method fitted on the train data. It is based on Bayesian optimisation, with its standard building blocks~\cite{CowenRivers2022}, i.e., \emph{an optimisation function}, \emph{a surrogate model} and \emph{an acquisition function}.

The surrogate model that BayCon uses is a Random Forest -- a regression ensemble model with 100 trees that approximates the optimization function. It aims to maximise the conditional probability of finding regions of the search space that are dense with valuable counterfactuals, while balancing exploration/exploitation trade-off, hence why the method is Bayesian. 
The acquisition function is the \emph{expected improvement} as defined by~\citet{mockus1978application}, with the exploration--exploitation coefficient set to $0.01$~\cite{Lizotte2008}. 
BayCon's optimisation function captures (i) the distance in the feature space, (ii) the distance in the prediction space, and (iii) the number of altered features, all scaled to the $[0, 1]$ range. The distance in the prediction space ($S_y$ score) measures the distance between the output class that the black-box model assigns to the candidate counterfactual ($x^\prime$) and the target output. In BayCon, the $S_y$ score for a supervised classification task is a ``hard score'', i.e., it is $1$ if the candidate counterfactual is assigned the target class and $0$ otherwise.

\subsection{Model-specific Soft Scoring}

To adapt BayCon to clustering algorithms we introduce soft output scores ($S_y$ score) that are suitable for clustering models.

\begin{figure}
    \centering
    \includegraphics[width=\linewidth]{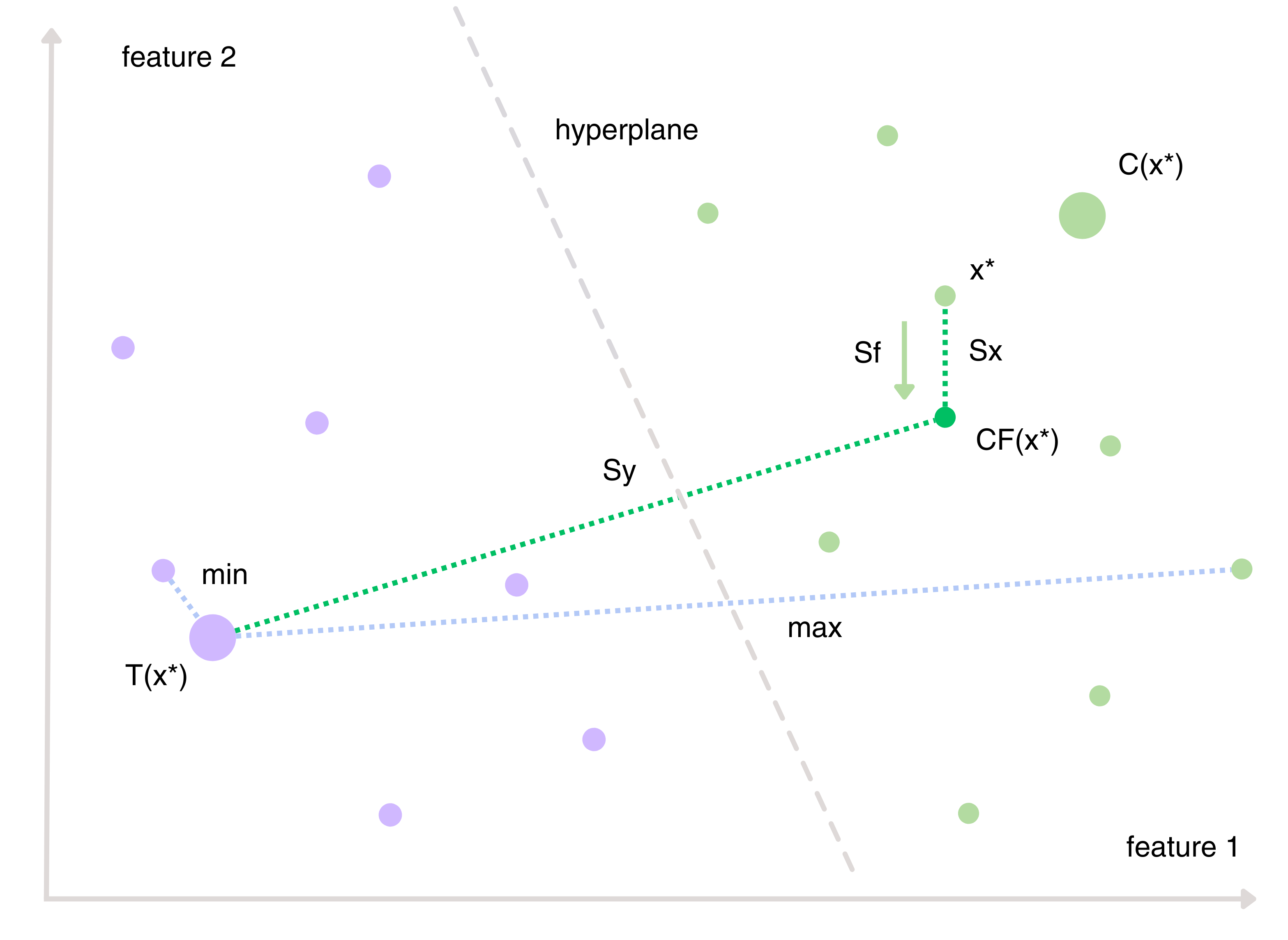}
    \caption[Example soft scores]{Representation of scores for the distance-based soft-scoring technique applied to a two-dimensional toy example. The target cluster is depicted in purple; large markers represent cluster centroids; and $x^\star$ is the initial instance. $S_x$ is the distance between $x^\star$ and the candidate $\mathit{CF}(x^\star)$; $S_f$ is the number of features changed (in this example $1$); and $S_y$ is the normalised distance to the centroid $C_{t}$ of the target cluster (see Equation~\ref{eq:score}).}
    \label{soft_k}
    \vspace*{6ex}
\end{figure}

\paragraph{Distance-Based Soft Scoring}
We exploit pairwise distances between the candidate counterfactuals and cluster summaries to speed up the counterfactual search and reduce score variance. To calculate this score (see Figure \ref{soft_k} for more information), we need (1) the centroid $C_{i}$ for each cluster $i$; (2) the minimum ($\min_{i}$) and the maximum ($\max_{i}$) distance between the centroid of the cluster and the corresponding data points in that cluster; and (3) the distance metric $d: \mathcal{X} \times \mathcal{X} \mapsto \mathbb{R}^+$ used during the clustering procedure (many clustering algorithms treat the distance metric as a hyperparameter). Given a target cluster $t$, an initial instance $x^\star$, and a candidate counterfactual $CF(x^\star)$, 
  the score $S_{y}$ is calculated according to Equation~\ref{eq:score}. The score $S_{y}$ is scaled to the interval $[0, 1]$ and indicates whether the candidate counterfactual is approaching the centroid of the target cluster. Values smaller than $0$ and larger than $1$ are clipped to $0$ and $1$ respectively. Such clipping may occur when we have to deal with new points (unavailable during the cluster-building phase), that are closer to the center than the \textit{min} or farther than the \textit{max} known values for that cluster. By default, the Euclidean distance is used, but it is possible to employ the same distance metric the black-box model utilises in the score calculations.

\begin{equation}
    S_{y}(CF(x^\star), t) = 1 - \frac{d(CF(x^\star), C_{t}) - min_{t}}{max_{t} - min_{t}}
    \label{eq:score}
\end{equation}

\paragraph{Density-Based Soft Scoring}
The efficient HDBSCAN~\cite{campello2013density} implementation (HDBSCAN$\star$~\cite{mcinnes2017hdbscan}) performs quick pre-computation during model fitting to speed up execution time. It determines where in the condensed tree the new data point would fall, assuming we do not change the condensed tree. Although the method is provided, the authors warn about the risk of drift in case unseen points are not cached, and the model is not retrained periodically. 
The soft clustering for HDBSCAN$\star$ leverages the condensed tree serving as a smoothed density function over data points and representative points to enable probabilistic clustering, i.e., the clustering model assigns each data point a vector of probabilities indicating its likelihood of belonging to a cluster. We utilise these probabilities as a distance metric in the output space.

\subsection{Model-agnostic Soft Scoring}
Our proposed model-agnostic soft-scoring method first extracts representative points for each cluster using prototypes and criticisms~\cite{Kim2016ExamplesAN}; next, the selected data points are used to train a semi-supervised self-training classifier that assigns membership probabilities based on the clusters. The membership probability corresponding to the target cluster can then be used as a soft-scoring metric.

\paragraph{Prototypes and Criticisms}
Prototypes and criticisms~\cite{Kim2016ExamplesAN} relies on the Maximum Mean Discrepancy (MMD), which is a discrepancy measure between two distributions~\cite{gretton2012kernel}. 
This approach -- called MMD-critic -- aims to minimise the divergence between the distribution of selected prototypes and the actual training data distribution. By leveraging kernel density estimation, the method approximates the data density and subsequently measures MMD. 
Mathematically, the MMD\textsuperscript{2} measure can be formulated as~\cite{molnar2022interpretable}
$$
\frac{1}{m^2} \sum_{i,j=1}^{m} k(z_i, z_j) - \frac{2}{mn} \sum_{i,j=1}^{m,n} k(z_i, x_j) + \frac{1}{n^2} \sum_{i,j=1}^{n} k(x_i, x_j)
\text{~.}
$$
Here $k$ represents the kernel function that measures the similarity between two points; $z$ and $x$ are respectively prototypes and data instances; $m$ denotes the number of prototypes; and $n$ signifies the number of data points. The choice of the kernel function is vital, with the radial basis function (RBF) kernel being a common pick because of its capacity to weigh points on a $0$--$1$ scale according to their proximity. 

The MMD-critic algorithm follows a systematic procedure for both prototype selection and criticism identification. The algorithm iteratively selects prototypes based on their MMD reduction potential, ensuring that the selected prototypes closely resemble the data distribution. Criticism identification relies on a witness function that quantifies the discrepancy between prototype and data density; it is formulated as 
$$
witness(x) = \frac{1}{n} \sum_{i=1}^{n} k(x, x_i) - \frac{1}{m} \sum_{j=1}^{m} k(x, z_j)
\text{~.}
$$
The witness function offers a mechanism to scrutinise instances where prototypes diverge from data instances, thereby serving as valuable counter-examples.

We adapted the MMD-critic implementation~\cite{Kim2016ExamplesAN} to our problem setting by selecting a fixed share of prototypes and criticisms for each cluster that the black-box model has discovered. All the parameters were kept default, including the 1:4 ratio between criticisms and prototypes. The fixed share of the percentage of representative points to feed to the self-training classifier was set to 20\% for each cluster. This share can be changed to account for data availability and the maximum number of points that can fit in memory for large datasets.

\paragraph{Semi-Supervised Learning}
Prototypes and criticisms, along with their cluster labels, are suitable for the initial training of a semi-supervised model.
Semi-supervised clustering capitalises on domain background knowledge to enhance conventional unsupervised clustering. This approach employs a limited number of prior information to predict cluster assignment for unlabeled data, resulting in a model capable of predicting new data. According to the taxonomy proposed by \citet{Cai2023ssc}, we are focusing on class labels (from the labelled prototypes and criticisms) as prior information. We chose a Self Training Classifier (STC)~\cite{yarowsky1995unsupervised}, a meta-estimator that offers a wide choice of base estimators. For our experiments,  we used the Extra Trees Classifier~\cite{geurts2006} as a base estimator. While it is an ensemble learning method similar to Random Forests, it comes with improvements that are pertinent to our case; specifically (i) increased randomness~\cite{geurts2006}, (ii) reduced variance~\cite{huang2005using}, (iii) faster training~\cite{pedregosa2011scikit}, and (iv) lower sensitivity to hyperparameters~\cite{geurts2012learning}. The STC outputs a vector of probabilities indicating the likelihood of belonging to every cluster for each data point. We utilise these probabilities as a distance metric in the output space.

\section{Experiments}

Given that there are no existing methods for generating counterfactuals for clustering models that could be used as a baseline, we compare our method against BayCon on two real-life datasets with two representative clustering algorithms.
To ensure traceability and reproducibility of our experiments, we mirror the setup described in the BayCon study~\cite{romashov2022baycon} (which does not need to be altered despite our work focusing on unsupervised rather than supervised learning).
Our method is implemented in Python 3.9 and relies on scikit-learn. All the experiments were run on a dual-core 1.1 GHz Intel Core i3 CPU with 8GB of RAM. We imposed a 15-minute runtime limit on each execution.

\subsection{Setup}

\paragraph{Datasets}
 
We experiment on five datasets: two with $k$-means++, two with HDBSCAN, and one with both such that each algorithm is tested on three datasets in total. Their summary characteristics are given in Table~\ref{tab:summary_data}.

For each dataset, we selected ten random instances to be explained, generating their counterfactuals three times to account for randomness, i.e., 30 runs per dataset.
We chose five initial instances to be explained from each of the top two clusters with the highest cardinality, i.e., starting from the initial instance selected to be explained that belongs to cluster \#1, the goal is to generate counterfactual instances (explanations) that belong to cluster \#2, and vice versa. For example, if we have three clusters of cardinality A:13, B:18, and C:15, we take 5 random instances from B and try to generate explanations belonging to C, and 5 from C to perform the opposite task. 

\begin{table}[t]
    \centering    
\begin{tabular}{@{}l@{\hskip 5pt}l@{\hskip 5pt}r@{\hskip 5pt}r@{\hskip 5pt}l@{}}

 \toprule
 Dataset & Domain & Feature \# & Instance \# & Model \\ 
 \midrule
 Diabetes & Medicine & 8 & 768 & $k$-means++ \\ 
 Biodeg & Biology & 41 & 1,055 & $k$-means++ \\ 
 Customer & Marketing & 25 & 2,216 & HDBSCAN \\ 
 Churn & Business & 19 & 1,000 & HDBSCAN \\ 
 Wine & Chemistry & 13 & 178 & Both \\ 
 
 \bottomrule
\end{tabular}
    \caption[Datasets summary]{Datasets summary.\label{tab:summary_data}}
    \vspace*{3ex}
\end{table}

\paragraph{Clustering models}

The explanations were generated for two clustering algorithms: $k$-means++ and HDBSCAN. Tables~\ref{tab:hdb} and \ref{tab:kmeans} present summary statistics for the clusters per dataset.


\begin{table}[b]
    \centering    
\begin{tabular}{@{}l@{\hskip 5pt}r@{\hskip 5pt}r@{\hskip 5pt}r@{}}

 \toprule
 Dataset & Min.\ Size & Cardinalities & Outliers \\ 
 \midrule
 Customer & 100 & [325, 146, 625] & 1120 (50\%) \\ 
 Churn & 20 & [497, 60, 74, 50] & 319 (32\%) \\ 
 Wine & 20 & [21, 119] & 38 (21\%) \\ 
 
 \bottomrule
\end{tabular}
    \caption[Datasets summary]{Summary of clusters for HDBSCAN.\label{tab:hdb}}
    \vspace*{3ex}
\end{table}

\begin{table}[b]
    \centering    
\begin{tabular}{@{}l@{\hskip 5pt}r@{\hskip 5pt}r@{}}

 \toprule
 Dataset & Cluster \# & Cardinalities\\ 
 \midrule
 Diabetes & 3 & [495, 235, 38] \\ 
 Biodeg & 3 & [407, 236, 412] \\ 
 Wine & 3 & [69, 47, 62] \\ 
 
 \bottomrule
\end{tabular}
    \caption[Datasets summary]{Summary of clusters for $k$-means++.\label{tab:kmeans}}
\end{table}
We used default parameter values with HDBSCAN except for \emph{minimum cluster size}, which we set to 15 as suggested by the method's documentation when dealing with datasets with heavy class imbalance or many suspected outliers (which HDBSCAN assigns to a dedicated cluster).
%
For $k$-means++ we also used default parameter values except for $k$, which was chosen experimentally by looking at t-SNE 2D visualisation and Silhouette scores.
The distance measure was kept as Euclidean, nevertheless our method is compatible with any distance metric for tabular data.
Testing the performance of the clustering methods is out of the scope of this study.
Note that clustering models are not used on all datasets as we selected for each model only the datasets that showed the best performance based on the t-SNE 2D visualisation, Silhouette scores (for $k$-means++) and condensed tree (for HDBSCAN). This selective approach allowed us to test each counterfactual generator on the most effectively clustered datasets, ensuring fair comparison and analysis.

\paragraph{Evaluation Metrics}
Mirroring the BayCon study, we used two types of evaluation metrics. The first one captures the quality of the generated counterfactuals, and the second pertains to the computational performance of the explainability algorithm. To quantify the quality of the generated counterfactuals, we computed \emph{similarity in the feature space} based on the Gower distance (Score $x$) and \emph{proportion of features} that were not tweaked (Score $f$). These are the same metrics used in the BayCon study, allowing a fair comparison with the baseline \cite{romashov2022baycon}.

We also used a third metric -- which we refer to as \emph{explained instances \%}, or \emph{Exp\%} in the result tables -- calculated as the percentage of trials in which at least one counterfactual explanation was found; e.g., if the initial instance is explained in 15 out of 30 trials, Exp\% is 50\%. It should be noted that this metric is analogous to the metric used by BayCon for classification tasks named \emph{Score $y$}, which captures similarity in the output space. However, we avoid naming it Score $y$ because all of the methods used in our experiments used a different Score $y$, i.e., different scoring in the output space.

%
To capture the computational performance of the explainers we measure \emph{time to first counterfactual} (in seconds), \emph{time to best counterfactual} (in seconds), and \emph{number of generated counterfactuals}. 
%
A core premise of BayCon was to generate high-scoring counterfactuals as fast as possible. Since our method requires additional calculation during the search phase, it is important to measure compute time. We report the results of each metric as the mean and standard deviation computed across all runs.

\subsection{Results}

\paragraph{Counterfactual Quality Results}
Table~\ref{tab:quality_km} reports the quality metrics for the baseline as well as the model-specific and model-agnostic variants of our method.
For the input (i.e., feature) space measures -- Score $x$ and Score $f$ -- we can observe that all methods achieve similar scores. This was expected given that our method focuses on the distance metric (soft scores) in the clustering output space. This means that the changes introduced by our novel soft-scoring methods (model-specific and model-agnostic) do not negatively influence the quality metrics related to the input space.
Regarding Exp\%, the \textit{model-agnostic} method improves upon the \textit{baseline}. It even outperforms the \textit{model-specific} methods in 2 (out of 6) dataset--black-box algorithm combinations and equates their performance in another 2 combinations.

More specifically, for \textbf{$k$-means++} the \textit{baseline} was the worst-performing method across all the datasets except Biodeg, where all methods achieved comparable scores. On Diabetes the \textit{model-agnostic} method achieved an increase of 33 percentage points in Exp\%, raising it from $\nicefrac{20}{30}$ (67\%) to $\nicefrac{27}{30}$ (90\%). On Wine the \textit{model-specific} method was the best performing overall, but \textit{model-agnostic} still outperformed the \textit{baseline} with a 17 percentage points increase in Exp\%.

For \textbf{HDBSCAN} the \textit{model-agnostic} method found counterfactuals in 100\% of the cases on the Customer and Wine datasets, a significant improvement compared to the \textit{baseline}, which explained 63\% of the instances on the former and 93\% on the latter dataset. For Biodeg, on the other hand, the \textit{model-specific} method was the best performing with 80\% explained instances; nonetheless, the \textit{model-agnostic} method outperformed the \textit{baseline} by 10 percentage points. Note how the \textit{baseline} may fail to find counterfactuals in some cases since it relies on a binary measure -- 1 for reaching the target and 0 otherwise, inherited from BayCon~\cite{romashov2022baycon,Dandl2020} -- which overlooks the continuous spatial information essential for guiding the counterfactual search. Without this information, the \textit{baseline} lacks the guidance necessary for finding counterfactuals when initial instances are isolated from the target clusters.

\begin{table}[t]
    \centering    
\begin{tabular}{@{}l@{\hskip 5pt}l@{\hskip 5pt}r@{\hskip 5pt}r@{\hskip 5pt}r@{\hskip 4pt}r@{}}

 \toprule
 Dataset & Method & Exp\% $\uparrow$ & Score $x$ $\uparrow$ & Score $f$ $\uparrow$ \\ 
 \midrule
 
 Diabetes & \textit{Baseline} & 67\% & \textbf{0.95(0.02)} & 0.76(0.02) \\ 
 (\emph{KM}) & \textit{Agnostic} & \textbf{90\%} & 0.94(0.03) & \textbf{0.78(0.3)} \\ 
 & \textit{Specific} & \textbf{90\%} & \textbf{0.95(0.03)} & 0.77(0.02) \\ 

 Biodeg & \textit{Baseline} & 100\% & 0.98(0.00) & 0.92(0.02) \\ 
 (\emph{KM}) & \textit{Agnostic} & 100\% & 0.98(0.00) & 0.92(0.01) \\ 
 & \textit{Specific} & 100\% & 0.98(0.00) & \textbf{0.93(0.01)} \\ 

 Wine & \textit{Baseline} & 77\% & 0.93(0.01) & 0.82(0.01) \\ 
 (\emph{KM}) & \textit{Agnostic} & 90\% & 0.93(0.01) & \textbf{0.83(0.02)} \\ 
 & \textit{Specific} & \textbf{100\%} & 0.93(0.01) & 0.82(0.01) \\ 

 Customer & \textit{Baseline} & 63\% & \textbf{0.97(0.01)} & \textbf{0.89(0.02)} \\ 
 (\emph{HC}) & \textit{Agnostic} & \textbf{100\%} & 0.96(0.01) & 0.88(0.02) \\ 
 & \textit{Specific} & 77\% & \textbf{0.97(0.01)} & 0.88(0.02) &\\ 

 Churn & \textit{Baseline} & 57\% & \textbf{0.97(0.01)} & 0.85(0.01) \\ 
 (\emph{HC}) & \textit{Agnostic} & 67\% & 0.96(0.01) & 0.85(0.02) \\ 
 & \textit{Specific} & \textbf{80\%} & 0.96(0.01) & 0.85(0.01) \\ 

 Wine & \textit{Baseline} & 93\% & 0.96(0.01) & 0.81(0.02) \\ 
 (\emph{HC}) & \textit{Agnostic} & \textbf{100\%} & 0.96(0.01) & \textbf{0.82(0.02)} \\ 
 & \textit{Specific} & 97\% & 0.90(0.01) & 0.81(0.02) \\ 
 
 \bottomrule
\end{tabular}
    \caption[Quality on $k$-means++]{Counterfactual quality measurements for $k$-means++ (\emph{KM}) and HDBSCAN (\emph{HS}).\label{tab:quality_km}}
    \vspace*{3ex}
\end{table}

\paragraph{Computational Performance Results}
The results of these experiments are presented in Table~\ref{tab:perf} and Table~\ref{tab:mixed}. The Wine dataset is analyzed in a separate table (Table~\ref{tab:mixed}) because for this dataset we we have both \textbf{HDBSCAN} and \textbf{$k$-means++} as clustering models. For the other datasets (Table~\ref{tab:perf}), we have only one clustering model per dataset.
%

Despite our concerns about higher computational complexity due to the increase in number of operations required by soft-scoring methods, they perform better on almost all the metrics for the two $k$-means-only datasets. Specifically for \textbf{$k$-means++} on the Diabetes dataset, the times to the first and best counterfactuals are roughly halved, respectively
from 3.24 for the \textit{baseline} to 1.52 for the \textit{model-agnostic} and 1.27 for the \textit{model-specific} methods, and
from 5.32 for the \textit{baseline} to 2.97 for the \textit{model-agnostic} and 2.18 for the \textit{model-specific} methods.
On the Biodeg dataset, the times to the first counterfactual are also rougly halved, from 1.11 for the \textit{baseline} to 0.63 for the soft-scoring method; the times to the best counterfactual remain largely the same, although the \textit{baseline} performed better (6.98 versus 7.12 for the \textit{model-specific} method). The number of counterfactuals is indecisive since on Diabetes the \textit{baseline} performed better and on Biodeg the \textit{model-specific} method generated more counterfactuals. 

In the \textbf{HDBSCAN} experiments, the improvement compared to the baseline is even more pronounced. For example, on the Customer dataset, the times recorded for the \textit{model-agnostic} method are one order of magnitude smaller than for the \textit{baseline}, with 3.85 against 20.96 for the time to first and 8.92 against 23.92 for the time to best. The \textit{model-specific} method for HDBSCAN generated slightly more counterfactuals, e.g., 97 on Churn and 139 on Customer; nonetheless, the \textit{model-agnostic} method generated 73 on Churn and 110 on Customer, proving that variety was not sacrificed. 

\begin{table}[t]
    \centering    
\begin{tabular}{@{}l@{\hskip 5pt}l@{\hskip 5pt}r@{\hskip 5pt}r@{\hskip 5pt}r@{}}
 \toprule
 Dataset & Method & \emph{t} to 1\textsuperscript{st} $\downarrow$ & \emph{t} to best\ $\downarrow$ & CF \# $\uparrow$ \\ 
 \midrule
 Diabetes & \textit{Baseline} & 3.24(7.27) & 5.32(7.06) & \textbf{278(355)} \\ 
 
 (\emph{KM}) & \textit{Agnostic} & 1.52(2.73) & 2.97(4.12) & 240(318) \\

 & \textit{Specific} & \textbf{1.27(1.56)} & \textbf{2.18(1.61)} & 253(314) \\ 

 Biodeg & \textit{Baseline} & 1.11(1.69) & \textbf{6.98(6.98)} & 880(588) \\ 
 
 (\emph{KM}) & \textit{Agnostic} & \textbf{0.63(0.91)} & 7.89(5.75) & 851(673) \\

 & \textit{Specific} & \textbf{0.63(0.61)} & 7.12(5.43) & \textbf{931(715)} \\ 

 Customer & \textit{Baseline} & 20.96(21.25) & 23.92(20.53) & 100(124) \\ 
 
 (\emph{HS}) & \textit{Agnostic} & \textbf{3.85(3.44)} & \textbf{8.92(6.43)} & 110(83) \\

 & \textit{Specific} & 13.31(15.36) & 19.56(18.76) & \textbf{139(127)} \\ 

 Churn & \textit{Baseline} & 8.05(14.26) & 9.08(14.42) & 96(99) \\ 
 
 (\emph{HS}) & \textit{Agnostic} & \textbf{5.31(10.53)} & \textbf{7.69(11.45)} & 73(65) \\

 & \textit{Specific} & 17.66(26.90) & 21.02(27.02) & \textbf{97(69)} \\ 
 
 \bottomrule
\end{tabular}
    \caption[Computational performance]{Computational performance for $k$-means++ (\emph{KM}) and HDBSCAN (\emph{HS}); time (denoted as \emph{t}) given in seconds.\label{tab:perf}}
    \vspace*{3ex}
\end{table}


The results for the Wine dataset are reported in  Table~\ref{tab:mixed}. The \textit{Baseline} generated the highest number of counterfactuals in both cases (203 for $k$-means++, 243 for HDBSCAN). The \textit{Model-specific} for $k$-means performed better on times, whereas on HDBSCAN, the \textit{baseline} performed better. Specifically, for $k$-means++, times to the first and best counterfactuals were slightly lower but of the same order of magnitude e. They went, respectively, from 2.69 of \textit{baseline} to 1.58 of \textit{model-specific} for $k$-means and from 3.58 of \textit{baseline} to 2.87 of \textit{model-specific} for $k$-means. On HDBSCAN, times to the first counterfactual are 0.66 for \textit{baseline} and above 1 for the soft-scoring methods. The times to the best counterfactual are 2.99 for \textit{baseline} and above 8 for the other two methods.

We can conclude that the goal of not sacrificing computational performance was still achieved. Additionally, in the majority of the experiments, times were even lower for the new versions.

\begin{table}[t]
    \centering    
\begin{tabular}{@{}l@{\hskip 5pt}l@{\hskip 5pt}r@{\hskip 5pt}r@{\hskip 5pt}r@{}}
 \toprule
 Model & Method & \emph{t} to 1\textsuperscript{st} $\downarrow$ & \emph{t} to best $\downarrow$ & CF \# $\uparrow$ \\ 
 \midrule
 \emph{KM} & \textit{Baseline} & 2.69(1.97) & 3.58(1.90) & \textbf{203(118)} \\ 
 
 & \textit{Agnostic} & 5.47(8.70) & 8.47(10.31) & 129(60) \\

 & \textit{Specific} & \textbf{1.58(0.69)} & \textbf{2.87(1.13)} & 129(21) \\ 

 \emph{HS} & \textit{Baseline} & \textbf{0.66(0.61)} & \textbf{2.99(2.82)} & \textbf{243(72)} \\ 
 
 & \textit{Agnostic} & 1.37(0.61) & 13.02(2.28) & 149(72) \\

 & \textit{Specific} & 5.12(13.97) & 8.12(15.25) & 238(70) \\ 
 
 \bottomrule
\end{tabular}
    \caption[Mixed experiments computational performance]{Computational performance on Wine for $k$-means++ (\emph{KM}) and HDBSCAN (\emph{HS}); time (denoted as \emph{t}) given in seconds.\label{tab:mixed}}
    \vspace*{3ex}
\end{table}

\section{Discussion}

The proposed model-specific soft-scoring methods for $k$-means++ and HDBSCAN, as well as the novel model-agnostic method, perform better than the (hard-scoring) baseline method across the board. Specifically, the model-agnostic method performed better than the baseline method in terms of the number of explained instances for all the datasets, and at times, it performed comparable to or better than the two model-specific methods. 
Additionally, the model-agnostic method found valid counterfactuals for each explained instance in at least 1 out of 3 generation attempts while the model-specific methods failed at times despite performing better overall. The computational performance and distance in the input space (Score $x$ in the result tables) were not sacrificed at all, while the number of features changed (and the associated Score $f$ in the result tables) was slightly better in most cases. The Mann-Whitney \textit{U} Test applied to each metric confirmed that there is no significant difference between the scores of the baseline and soft methods.
This result suggests that although more computation is necessary for the soft-scoring method, exploiting the spatial information allows us to better guide the search and speed it up. Overall counterfactual multiplicity and quality are preserved while more instances are explained, indicating that searching for solutions with soft scores where the baseline failed to find any does not lead to regions with low-quality alternatives. 

\paragraph{$k$-means++}
In the experiments involving $k$-means++, the \textit{model-specific} method always performed at least as well as the \textit{model-agnostic} method. This result was expected since an ad-hoc scaled distance-based method is likely to be the best soft-scoring technique, given that it is specific to the given model. Regardless, the difference in \emph{Explained instances \%} (\emph{Exp\%} in the result tables) was 0\% on the Diabetes and just 10\% on the Wine dataset. Score $x$ was the same (with the chosen rounding) in almost all trials; Score $f$, although less regular, showed negligible differences. Overall, the best improvement was obtained on the Diabetes dataset, where the number of explained instances increased from 67\% for the \textit{baseline} to 90\% for both the \textit{model-specific} and \textit{model-agnostic} methods.

These results imply that the self-training classifier initiated with a fixed percentage of prototypes and criticisms for each cluster (used by the \textit{model-agnostic} approach) can uncover patterns in the cluster assignment without explicitly relying on centroid distance. While there might exist other formulations of the model-specific soft scoring that could improve the share of explained instances, we did not pursue this line of research given that Exp\% was already 90\% on Diabetes and 100\% on the other two datasets, 
thus any improvement would be marginal and irrelevant to our study.

\paragraph{HDBSCAN}
In the experiments involving HDBSCAN, the \textit{model-agnostic} method explained instances in \emph{all} experimental trials on two datasets -- Customer and Wine -- which is an improvement compared to both the \textit{baseline} and \textit{model-specific} methods. On Churn, however, the percentage of explained instances was lower across all methods. Notably, on this dataset HDBSCAN produced four clusters with one of them being 7 to 10 times larger than the other three in cardinality (see Table~\ref{tab:hdb}). This cluster imbalance may be the reason for slightly worse performance. 

\paragraph{Ablation}
We also investigated whether the \textit{model-agnostic} method is robust to hyperparameter changes to understand if its performance is stable~\cite{mania2018simple}. Specifically, we examined the share of initial representative points, which depends on the dataset size, available memory and privacy constraints. Intuition suggests that the larger the share, the better the results.
%
%
We set the percentage of representative points (which are used to train the self-training classifier) to 20\% in our initial experiments. We vary this value -- setting it to 10\% and 5\% -- to investigate its impact on the quality of counterfactuals output by the \textit{model-agnostic} method; the experiment setup remains otherwise unchanged. Table \ref{tab:kmeans_ssc}, presents the results for these experiments. 

\begin{table}[]
    \centering    
\begin{tabular}{@{}l@{\hskip 5pt}r@{\hskip 5pt}r@{\hskip 5pt}r@{\hskip 5pt}r@{\hskip 5pt}r@{}}

 \toprule
 Dataset & Rep\% & Exp\% $\uparrow$ & Score $x$ $\uparrow$ & Score $f$ $\uparrow$ \\ 
 \midrule
 
 Diabetes & 20 & \textbf{90\%} & 0.94(0.03) & \textbf{0.78(0.03)} \\ 
 (\emph{KM}) & 10 & 87\% & \textbf{0.95(0.02)} & 0.77(0.03) \\ 
 & 5 & 87\% & \textbf{0.95(0.03)} & 0.77(0.03) \\ 

 Biodeg & 20 & 100\% & 0.98(0.00) & 0.92(0.01) \\ 
 (\emph{KM}) & 10 & 100\% & 0.98(0.00) & 0.92(0.01) \\ 
 & 5 & 100\% & 0.98(0.00) & 0.92(0.01) \\ 

 Wine & 20 & 90\% & 0.93(0.01) & \textbf{0.83(0.02)} \\ 
 (\emph{KM}) & 10 & \textbf{93\%} & 0.93(0.01) & 0.82(0.02) \\ 
 & 5 & \textbf{93\%} & 0.93(0.01) & 0.82(0.02) \\ 

 Customer & 20 & \textbf{100\%} & 0.96(0.01) & 0.88(0.02) \\ 
 (\emph{HS}) &10 & 90\% & \textbf{0.97(0.01)} & \textbf{0.90(0.02)} \\ 
 &5 & 87\% & \textbf{0.97(0.01)} & 0.89(0.03) \\ 

 Churn & 20 & 67\% & 0.96(0.01) & 0.85(0.02) \\ 
 (\emph{HS}) & 10 & \textbf{70\%} & 0.96(0.01) & 0.85(0.01) \\ 
 & 5 & 67\% & 0.96(0.01) & 0.85(0.01) \\ 

 Wine & 20 & 100\% & 0.96(0.01) & 0.82(0.03) \\ 
 (\emph{HS}) & 10 & 100\% & 0.96(0.01) & \textbf{0.83(0.03)} \\ 
 & 5 & 100\% & 0.96(0.01) & \textbf{0.83(0.03)} \\ %
 
 \bottomrule
\end{tabular}
    \caption[\textit{Model-Agnostic} variations on $k$-means++]{Counterfactual quality measurements on $k$-means++ (\emph{KM}) and HDBSCAN (\emph{HS}) with representative points variations on the model-agnostic version.\label{tab:kmeans_ssc}}
    \vspace*{3ex}
\end{table}

On the Diabetes dataset, the ``20\%'' version performed slightly better across most metrics, with an instance more for which counterfactuals were generated, as shown by Exp\%. Conversely, on the Wine dataset, both lower percentages allowed to explain one instance more, although Score $f$ was marginally better for the larger share. On Biodeg, there were no relevant changes in the three Scores, except for some rounding in Score $f$.

While on the Customer dataset results are dramatically better, with a full 100\% in Exp\%, on Wine there are no significant changes. Interestingly, on Churn version ``10\%'' was the best-performing, with one instance more explained compared to both alternative versions. For the comparison with \textit{baseline}, \textit{model-specific} for $k$-means, and \textit{model-specific} for HDBSCAN, we kept the results with ``20\%'' representative points, although it was not the best performing in some cases. It should, however, be noted that the three \textit{model-agnostic} variants outperform \textit{baseline} in all metrics across the six method-dataset pairs and, in some cases, \textit{model-specific} for $k$-means and \textit{model-specific} for HDBSCAN as well.

In general, no other setting led to significantly better results, with only marginal improvements observed if at all, which suggest that the method is robust. For example, in the best-case scenario (for the Customer dataset) the cardinality of the selected data points has to be doubled (e.g., from 10\% to 20\%) to improve the percentage of explained instances by 10 percentage points (from 90\% to 100\%). While being able to explain all instances is promising, more experiments on different datasets are needed to verify this finding.
Similarly to the initial experiments, the soft-scoring method achieved comparable scores to those achieved by the baselines in quality measures related to the input (i.e., feature) space (Score $x$ and Score $f$ in Table \ref{tab:kmeans_ssc}). This observation is an additional confirmation that the changes introduced by our novel soft-scoring method (which targets the output space) did not negatively impact the quality metrics related to the input space.
%

\paragraph{Limitations} One limitation of our study is the absence of a user study. Without user feedback on generated counterfactuals, it is difficult to validate the practical effectiveness of the generated counterfactual explanations. Additionally, our approach does not include a grid-search hyperparameter tuning. For this reason, the experiments might underestimate the quality of generated counterfactual explanations. Furthemore, the inclusion of other types of data such as images, text, or audio might require clustering unstructured data in an actionable feature space before utalizing the proposed method.

\section{Conclusion}

To the best of our knowledge, this is the first XAI method based on counterfactual explanations specifically designed to explain clustering assignments and, in general, one of the few post-hoc methods addressing unsupervised learning. 
The experiments, performed on five datasets and two clustering algorithms, show that introducing soft scores to guide counterfactual search improves the performance compared to a state-of-the-art counterfactual generator. These results encourage exploring further refinements, like tweaking other components of the method to produce counterfactuals of even higher quality and, at the same time, addressing counterfactual multiplicity to provide just a selection of meaningful explanations~\cite{sokol2023navigating}. %


\bibliography{mybibfile}

\begin{thebibliography}{40}
\providecommand{\natexlab}[1]{#1}
\providecommand{\url}[1]{\texttt{#1}}
\expandafter\ifx\csname urlstyle\endcsname\relax
  \providecommand{\doi}[1]{doi: #1}\else
  \providecommand{\doi}{doi: \begingroup \urlstyle{rm}\Url}\fi

\bibitem[Agrawal et~al.(1998)Agrawal, Gehrke, Gunopulos, and Raghavan]{agrawal1998automatic}
R.~Agrawal, J.~Gehrke, D.~Gunopulos, and P.~Raghavan.
\newblock Automatic subspace clustering of high dimensional data for data mining applications.
\newblock In \emph{Proceedings of the 1998 ACM SIGMOD international conference on Management of data}, pages 94--105. Association for Computing Machinery, 1998.

\bibitem[Ali et~al.(2023)Ali, Abuhmed, El-Sappagh, Muhammad, Alonso-Moral, Confalonieri, Guidotti, {Del Ser}, Díaz-Rodríguez, and Herrera]{Ali2023}
S.~Ali, T.~Abuhmed, S.~El-Sappagh, K.~Muhammad, J.~M. Alonso-Moral, R.~Confalonieri, R.~Guidotti, J.~{Del Ser}, N.~Díaz-Rodríguez, and F.~Herrera.
\newblock Explainable artificial intelligence ({XAI}): What we know and what is left to attain trustworthy artificial intelligence.
\newblock \emph{Information Fusion}, 2023.

\bibitem[Altmann et~al.(2010)Altmann, Toloşi, Sander, and Lengauer]{Altmann2010}
A.~Altmann, L.~Toloşi, O.~Sander, and T.~Lengauer.
\newblock Permutation importance: a corrected feature importance measure.
\newblock \emph{Bioinformatics}, 2010.

\bibitem[Cai et~al.(2023)Cai, Hao, Yang, Zhao, and Yang]{Cai2023ssc}
J.~Cai, J.~Hao, H.~Yang, X.~Zhao, and Y.~Yang.
\newblock A review on semi-supervised clustering.
\newblock \emph{Information Sciences}, 632:\penalty0 164--200, 2023.

\bibitem[Campello et~al.(2013)Campello, Moulavi, and Sander]{campello2013density}
R.~J. Campello, D.~Moulavi, and J.~Sander.
\newblock Density-based clustering based on hierarchical density estimates.
\newblock In \emph{Advances in Knowledge Discovery and Data Mining: 17th Pacific-Asia Conference, PAKDD 2013, Gold Coast, Australia, April 14-17, 2013, Proceedings, Part II 17}, pages 160--172. Springer, 2013.

\bibitem[Cirrincione et~al.(2020)Cirrincione, Ciravegna, Barbiero, Randazzo, and Pasero]{cirrincione2020gh}
G.~Cirrincione, G.~Ciravegna, P.~Barbiero, V.~Randazzo, and E.~Pasero.
\newblock The {GH-EXIN} neural network for hierarchical clustering.
\newblock \emph{Neural Networks}, 121:\penalty0 57--73, 2020.

\bibitem[Cowen-Rivers et~al.(2022)Cowen-Rivers, Lyu, Tutunov, Wang, Grosnit, Griffiths, Jianye, Wang, Peters, and Ammar]{CowenRivers2022}
A.~I. Cowen-Rivers, W.~Lyu, R.~Tutunov, Z.~Wang, A.~Grosnit, R.-R. Griffiths, H.~Jianye, J.~Wang, J.~Peters, and H.~Ammar.
\newblock {HEBO}: An empirical study of assumptions in {Bayesian} optimisation.
\newblock \emph{Journal of Artificial Intelligence Research}, 2022.

\bibitem[Dandl et~al.(2020)Dandl, Molnar, Binder, and Bischl]{Dandl2020}
S.~Dandl, C.~Molnar, M.~Binder, and B.~Bischl.
\newblock Multi-objective counterfactual explanations.
\newblock In T.~B{\"a}ck, M.~Preuss, A.~Deutz, H.~Wang, C.~Doerr, M.~Emmerich, and H.~Trautmann, editors, \emph{Parallel Problem Solving from Nature -- PPSN XVI}. Springer, 2020.

\bibitem[Ellis et~al.(2021)Ellis, Sendi, Geenjaar, Plis, Miller, and Calhoun]{ellis2021algorithm}
C.~A. Ellis, M.~S. Sendi, E.~Geenjaar, S.~M. Plis, R.~L. Miller, and V.~D. Calhoun.
\newblock Algorithm-agnostic explainability for unsupervised clustering.
\newblock \emph{arXiv preprint arXiv:2105.08053}, 2021.

\bibitem[Ester et~al.(1996)Ester, Kriegel, Sander, and Xu]{ester1996density}
M.~Ester, H.-P. Kriegel, J.~Sander, and X.~Xu.
\newblock A density-based algorithm for discovering clusters in large spatial databases with noise.
\newblock In \emph{Proceedings of the Second International Conference on Knowledge Discovery and Data Mining}, pages 226--231. AAAI Press, 1996.

\bibitem[Ezugwu et~al.(2022)Ezugwu, Ikotun, Oyelade, Abualigah, Agushaka, Eke, and Akinyelu]{ezugwu2022comprehensive}
A.~E. Ezugwu, A.~M. Ikotun, O.~O. Oyelade, L.~Abualigah, J.~O. Agushaka, C.~I. Eke, and A.~A. Akinyelu.
\newblock A comprehensive survey of clustering algorithms: State-of-the-art machine learning applications, taxonomy, challenges, and future research prospects.
\newblock \emph{Engineering Applications of Artificial Intelligence}, 110:\penalty0 104743, 2022.

\bibitem[Fahad et~al.(2014)Fahad, Alshatri, Tari, Alamri, Khalil, Zomaya, Foufou, and Bouras]{Fahad2014}
A.~Fahad, N.~Alshatri, Z.~Tari, A.~Alamri, I.~Khalil, A.~Y. Zomaya, S.~Foufou, and A.~Bouras.
\newblock A survey of clustering algorithms for big data: Taxonomy and empirical analysis.
\newblock \emph{IEEE Transactions on Emerging Topics in Computing}, 2014.

\bibitem[Gajcin(2023)]{gajcin2023rl}
J.~Gajcin.
\newblock Counterfactual explanations for reinforcement learning agents.
\newblock In \emph{Proceedings of the 2023 International Conference on Autonomous Agents and Multiagent Systems}. International Foundation for Autonomous Agents and Multiagent Systems, 2023.

\bibitem[Geurts and Louppe(2012)]{geurts2012learning}
P.~Geurts and G.~Louppe.
\newblock Learning to rank with extremely randomized trees.
\newblock \emph{Journal of Machine Learning Research}, 14:\penalty0 2753--2790, 2012.

\bibitem[Geurts et~al.(2006)Geurts, Ernst, and Wehenkel]{geurts2006}
P.~Geurts, D.~Ernst, and L.~Wehenkel.
\newblock Extremely randomized trees.
\newblock \emph{Machine Learning}, 2006.

\bibitem[Gretton et~al.(2012)Gretton, Borgwardt, Rasch, Sch{\"o}lkopf, and Smola]{gretton2012kernel}
A.~Gretton, K.~M. Borgwardt, M.~J. Rasch, B.~Sch{\"o}lkopf, and A.~Smola.
\newblock A kernel two-sample test.
\newblock \emph{The Journal of Machine Learning Research}, 13\penalty0 (1):\penalty0 723--773, 2012.

\bibitem[Huang and Ling(2005)]{huang2005using}
S.~Huang and C.~X. Ling.
\newblock Using {AUC} and accuracy in evaluating learning algorithms.
\newblock \emph{IEEE Transactions on Knowledge and Data Engineering}, 17\penalty0 (3):\penalty0 299--310, 2005.

\bibitem[Jain et~al.(1999)Jain, Murty, and Flynn]{Jain1999}
A.~K. Jain, M.~N. Murty, and P.~J. Flynn.
\newblock Data clustering: A review.
\newblock \emph{ACM Computing Surveys}, 31, 1999.

\bibitem[Kim et~al.(2016)Kim, Koyejo, and Khanna]{Kim2016ExamplesAN}
B.~Kim, O.~Koyejo, and R.~Khanna.
\newblock Examples are not enough, learn to criticize! {Criticism} for interpretability.
\newblock In \emph{Proceedings of the 30th International Conference on Neural Information Processing Systems}. Curran Associates Inc., 2016.

\bibitem[Lakkaraju et~al.(2016)Lakkaraju, Bach, and Leskovec]{Lakkaraju2016}
H.~Lakkaraju, S.~H. Bach, and J.~Leskovec.
\newblock Interpretable decision sets: A joint framework for description and prediction.
\newblock In \emph{Proceedings of the 22nd ACM SIGKDD International Conference on Knowledge Discovery and Data Mining}. Association for Computing Machinery, 2016.

\bibitem[Lizotte(2008)]{Lizotte2008}
D.~J. Lizotte.
\newblock \emph{Practical Bayesian Optimization}.
\newblock PhD thesis, University of Alberta, 2008.

\bibitem[Louhichi et~al.(2023)Louhichi, Nesmaoui, Mbarek, and Lazaar]{Louhichi2023}
M.~Louhichi, R.~Nesmaoui, M.~Mbarek, and M.~Lazaar.
\newblock Shapley values for explaining the black box nature of machine learning model clustering.
\newblock \emph{Procedia Computer Science}, 2023.

\bibitem[Lundberg and Lee(2017)]{Lundberg2017}
S.~M. Lundberg and S.-I. Lee.
\newblock A unified approach to interpreting model predictions.
\newblock In \emph{Proceedings of the 31st International Conference on Neural Information Processing Systems}. Curran Associates Inc., 2017.

\bibitem[Mania et~al.(2018)Mania, Guy, and Recht]{mania2018simple}
H.~Mania, A.~Guy, and B.~Recht.
\newblock Simple random search provides a competitive approach to reinforcement learning.
\newblock \emph{Advances in Neural Information Processing Systems}, 31, 2018.

\bibitem[McInnes and Healy(2017)]{mcinnes2017hdbscan}
L.~McInnes and J.~Healy.
\newblock Accelerated hierarchical density-based spatial clustering.
\newblock \emph{2017 IEEE International Conference on Data Mining Workshops (ICDMW)}, pages 33--42, 2017.

\bibitem[Mockus(1978)]{mockus1978application}
J.~Mockus.
\newblock Application of {Bayesian} approach to numerical methods of global and stochastic optimization.
\newblock \emph{Journal of Global Optimization}, 2\penalty0 (4):\penalty0 347--365, 1978.

\bibitem[Molnar(2022)]{molnar2022interpretable}
C.~Molnar.
\newblock \emph{Interpretable Machine Learning}.
\newblock 2nd edition, 2022.

\bibitem[Olson et~al.(2021)Olson, Khanna, Neal, Li, and Wong]{olson2021counterfactual}
M.~L. Olson, R.~Khanna, L.~Neal, F.~Li, and W.-K. Wong.
\newblock Counterfactual state explanations for reinforcement learning agents via generative deep learning.
\newblock \emph{Artificial Intelligence}, 295, 2021.

\bibitem[Pedregosa et~al.(2011)Pedregosa, Varoquaux, Gramfort, Michel, Thirion, Grisel, Blondel, Prettenhofer, Weiss, Dubourg, Vanderplas, Passos, Cournapeau, Brucher, Perrot, and Duchesnay]{pedregosa2011scikit}
F.~Pedregosa, G.~Varoquaux, A.~Gramfort, V.~Michel, B.~Thirion, O.~Grisel, M.~Blondel, P.~Prettenhofer, R.~Weiss, V.~Dubourg, J.~Vanderplas, A.~Passos, D.~Cournapeau, M.~Brucher, M.~Perrot, and E.~Duchesnay.
\newblock {scikit-learn}: {Machine} learning in {Python}.
\newblock \emph{Journal of Machine Learning Research}, 2011.

\bibitem[Poyiadzi et~al.(2020)Poyiadzi, Sokol, Santos-Rodriguez, De~Bie, and Flach]{poyiadzi2020face}
R.~Poyiadzi, K.~Sokol, R.~Santos-Rodriguez, T.~De~Bie, and P.~Flach.
\newblock {FACE}: {Feasible} and actionable counterfactual explanations.
\newblock In \emph{Proceedings of the AAAI/ACM Conference on AI, Ethics, and Society}, pages 344--350, 2020.

\bibitem[Ribeiro et~al.(2016)Ribeiro, Singh, and Guestrin]{RibeiroSingh2016}
M.~T. Ribeiro, S.~Singh, and C.~Guestrin.
\newblock ``{Why} should {I} trust you?": {Explaining} the predictions of any classifier.
\newblock In \emph{Proceedings of the 22nd ACM SIGKDD International Conference on Knowledge Discovery and Data Mining}. Association for Computing Machinery, 2016.

\bibitem[Romashov et~al.(2022)Romashov, Gjoreski, Sokol, Martinez, and Langheinrich]{romashov2022baycon}
P.~Romashov, M.~Gjoreski, K.~Sokol, M.~V. Martinez, and M.~Langheinrich.
\newblock {BayCon}: Model-agnostic bayesian counterfactual generator.
\newblock In \emph{Proceedings of the 31st International Joint Conference on Artificial Intelligence, Vienna, Austria}, pages 23--29, 2022.

\bibitem[Rudin(2019)]{rudin2019stop}
C.~Rudin.
\newblock Stop explaining black box machine learning models for high stakes decisions and use interpretable models instead.
\newblock \emph{Nature machine intelligence}, 1\penalty0 (5):\penalty0 206--215, 2019.

\bibitem[Scholbeck et~al.(2023)Scholbeck, Funk, and Casalicchio]{scholbeck2023algorithm}
C.~A. Scholbeck, H.~Funk, and G.~Casalicchio.
\newblock Algorithm-agnostic feature attributions for clustering.
\newblock In \emph{World Conference on Explainable Artificial Intelligence}, pages 217--240. Springer, 2023.

\bibitem[Sokol and Flach(2021)]{sokol2021explainability}
K.~Sokol and P.~Flach.
\newblock Explainability is in the mind of the beholder: Establishing the foundations of explainable artificial intelligence.
\newblock \emph{arXiv}, 2112.14466, 2021.

\bibitem[Sokol et~al.(2023)Sokol, Small, and Xuan]{sokol2023navigating}
K.~Sokol, E.~Small, and Y.~Xuan.
\newblock Navigating explanatory multiverse through counterfactual path geometry.
\newblock \emph{Workshop on Counterfactuals in Minds and Machines, ICML}, 2023.

\bibitem[Wang et~al.(1997)Wang, Yang, Muntz, et~al.]{wang1997sting}
W.~Wang, J.~Yang, R.~Muntz, et~al.
\newblock Sting: A statistical information grid approach to spatial data mining.
\newblock In \emph{Proceedings of the 23rd International Conference on Very Large Data Bases}, volume~97, pages 186--195. Morgan Kaufmann organizations Inc., 1997.

\bibitem[Xu and Tian(2015)]{Xu2015}
D.~Xu and Y.~Tian.
\newblock A comprehensive survey of clustering algorithms.
\newblock \emph{Annals of Data Science}, 3, 2015.

\bibitem[Yarowsky(1995)]{yarowsky1995unsupervised}
D.~Yarowsky.
\newblock Unsupervised word sense disambiguation rivaling supervised methods.
\newblock In \emph{Proceedings of the 33rd Annual Meeting on Association for Computational Linguistics (ACL '95)}. Association for Computational Linguistics, 1995.

\bibitem[Zhou et~al.(2022)Zhou, Xu, Zheng, Chen, Bu, Wu, Wang, Zhu, Ester, et~al.]{zhou2022comprehensive}
S.~Zhou, H.~Xu, Z.~Zheng, J.~Chen, J.~Bu, J.~Wu, X.~Wang, W.~Zhu, M.~Ester, et~al.
\newblock A comprehensive survey on deep clustering: Taxonomy, challenges, and future directions.
\newblock \emph{arXiv}, 2206.07579, 2022.

\end{thebibliography}

\end{document}